\documentclass[twoside]{article}
\usepackage{PRIMEarxiv}

\usepackage{times}
\usepackage{soul}
\usepackage{url}
\usepackage[hidelinks]{hyperref}
\usepackage[utf8]{inputenc}
\usepackage[small]{caption}
\usepackage{graphicx}
\usepackage{amsmath}
\usepackage{amsthm}
\usepackage{booktabs}
\usepackage[switch]{lineno}

\usepackage{caption} 
\usepackage{xspace}
\usepackage{amsmath}
\usepackage{macros}

\usepackage{svg}
\usepackage{tablefootnote}

\definecolor{darkgreen}{rgb}{0.0, 0.2, 0.13}

\title{Combining \CP Reasoning with Large Language Model Predictions}

\author{
    Florian Régin \\
    Université Côte d’Azur, CNRS, I3S, France\\
    florian.regin06@gmail.com \\
    \and
    Elisabetta De Maria \\
    Université Côte d’Azur, CNRS, I3S, France \\
    elisabetta.demaria@univ-cotedazur.fr
    \and
    Alexandre Bonlarron \\
    Université Côte d’Azur, Inria, France \\
    alexandre.bonlarron@gmail.com
}
\date{} % Leave this blank or specify a date

\begin{document}

\maketitle

\begin{abstract}

Constraint Programming (CP) and Machine Learning (ML) face challenges in text generation due to CP's struggle with implementing ``meaning'' and ML's difficulty with structural constraints. This paper proposes a solution by combining both approaches and embedding a Large Language Model (LLM) in CP. The LLM handles word generation and meaning, while CP manages structural constraints. This approach builds on 
GenCP, an improved version of On-the-fly Constraint Programming Search (OTFS) using LLM-generated domains.
Compared to Beam Search (BS), a standard NLP method, this combined approach
(GenCP with LLM)
is faster and produces better results, ensuring all constraints are satisfied. This fusion of CP and ML presents new possibilities for enhancing text generation under constraints.

\end{abstract}

\section{Introduction}
\label{sec:Intro}

 How can we perceive Constraint Programming beyond its traditional role in solving combinatorial optimization problems?
Once Eugene Freuder wrote \emph{
Constraint programming represents one of the closest approaches computer science has yet made to the Holy Grail of programming: the user states the problem, the computer solves it }\cite{freuder:97}.

Nevertheless, some real-world problems are still beyond the reach of the current CP paradigm. This is particularly true when real-world problems involve vague notions such as ``meaning'' and ``melody'' for text and music. %
These are not easy to model in CP with the classical toolbox, mainly because these notions are hard to define formally. For instance, it is unclear how to formalize an objective function or a constraint to get closer to a meaningful sentence, a melodious song or a captivating painting. 
On the other hand, recent results in Machine Learning (ML), such as transformer-based models \cite{vasmani-et-al:2017}, have demonstrated the power of these techniques to capture a significant part of these vague concepts through data-driven statistical learning (e.g., Large Language Model (LLM) like the GPT series \cite{radford:2019}, stable-diffusion \cite{rombach2021highresolution}, ChatMusician \cite{yuan2024chatmusician}). In the article, we demonstrate that ML, and in particular LLM, can help CP to model and solve problems where such vague concepts can be found.  

In recent years, there has been a growing interest in text generation under constraints thanks to the rise of transformer-based models, like OpenAI ChatGPT (\cite{radford:2019}) and Meta LLaMa (\cite{touvron2023llama}). Nevertheless, even fine-tuned 
prompted LLMs fail to generate several constrained outputs (see the tasks introduced in \cite{yao2023collie}).
The goal of this paper is to present a new method for the task of \tg. This interest has a strong chance of continuing to grow insofar as many brands wish to integrate these technologies, in particular with their customers, and want to have control and guarantees on the behavior of these conversational agents. Hence, it may impact several critical marketing aspects (e.g., brand representation, legal issues, data privacy, \dots). Therefore, CP has the potential to become a strong safeguard of this kind of generative model. 

For the task of \tg, ML techniques face limitations when they have to manage structural constraints, such as limits on the number of words or characters (e.g. Text Summarization, Text Simplification, Text style transfer, Question Answering, Storytelling, Poetry or Lyrics Generation, Subtitle) \cite{garbacea-arxiv-survey-nlp:2022}. CP succeeds on these types of constraints, making the combination of CP and ML a natural fit for the task of \tg. 

\begin{figure}[htbp]
    \centering
    \includegraphics[width=0.80\textwidth]{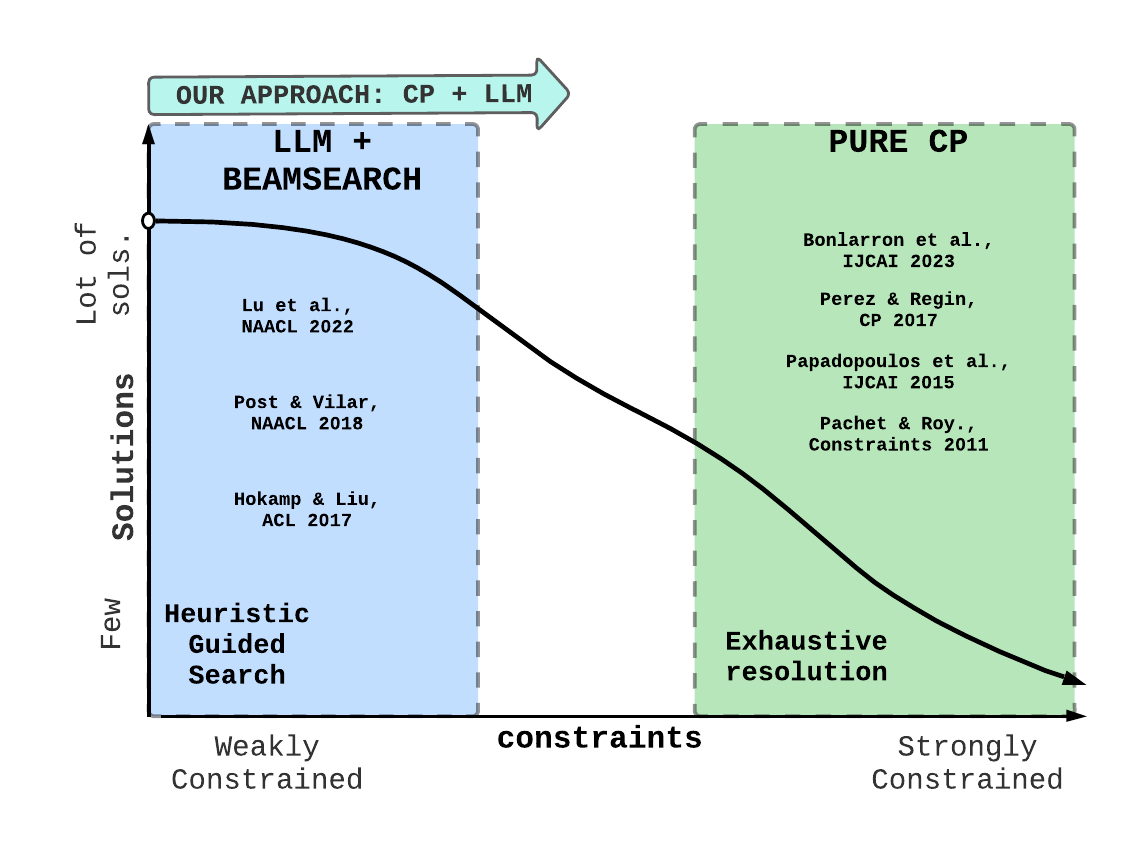}
    \caption{
    Our approach aspires to explore the in-between area. In the blue (left-hand side) region, LLM guided searches solve weakly constrained problems \cite{lu-etal-2022-neurologic,beamsearch2:2018,hokamp-liu:2017} and in the green (right-hand side) region, CP-based generation tackles strongly constrained problems \cite{sprockeels-vanroy:2024expressing,bonlarron-regin:2024inter,bonlarron-regin:2024markov,bonlarron-et-al:2023,perez-regin:17b,papadopoulos-roy-etal:15,pachet-roy:2011}.
    }
    \label{fig:enter-label}
\end{figure}

This paper proposes such a combination, to tackle a class of problems where neither CP and ML succeeds on their own (Fig.~\ref{fig:enter-label}).

 Combining Combinatorial Optimization (CO) and ML is a very active research area \cite{BENGIO-survey:2021}, however there is no easy way to integrate the ML ``expertise'' into CP as a constraint of the model \cite{bartolini-et-al-neuron:2011,Lombardi-et-al-EML:2017} and \emph{vice versa} \cite{willem}. Furthermore, there are many incentives to strengthen the interactions between ML and CO \cite{Khalil_et_al_mip:2016,Khalil-et-al-mip:2017,Song-et-al-LNS-MIP:2020}. Usually, the main motivation comes from the performance perspective, where the idea is to improve a solver's performance with ML (e.g., finding branching heuristics thanks to Reinforcement Learning \cite{Cappart-et-al:2021} or finding better  bounds with Clustering \cite{Nafar-Romer:2024}). This paper tackles it from the modeling point of view. Modeling is a crucial part of CO works. In the end, the model must account for the underlying solver that runs it. More in detail, here, the paper focuses on the interaction between CP and ML, more precisely through an ML-augmented CP approach \cite{kotary-survey:2021}. 

In the context of \tg, the domain of a variable represents a word. The base idea of the paper consists in letting ML manage the domain of variables and CP manage the constraints and the number of variables. In this manner, the sentence formed by variables has high chances to have a meaning and all the constraints will be satisfied. In traditional CP, the domains can not be managed by  ML because they have to be set beforehand. However, it is possible to rely on \OTF (\otf) \cite{FGCSPICTAI}, a CP based method where variables, domains and constraints are generated during the search for a solution. \color{black}

The main contribution of this paper is to propose a new version of \otf,
called GenCP,
where the generative function of the domain of variables is modified to allow CP variable domains to be computed by an LLM embedded in it, during the search for a solution. % 
More in detail, ML is used during  process solving but it is also used as an explicit part of the problem definition (i.e., domains are predicted by the LLM and can replace entirely static variable domains definition of a CSP.). Thus it bridges CP and ML through solving and modelling.
% Helping functions were introduced to separate implicit and explicit constraints: explicit constraints are constraints of the problem (a word does not contain an "e" for example), while implicit constraint represent all the other constraints, like ensuring that a variable corresponds to a word, not the beginning of a word etc$\dots$

The potential of the approach is showcased for the problem of \tg, against one the most used techniques in the Natural Language Processing (NLP) field: Beam Search (BS). Both methods 
% (BS and the new version of \otf) 
(BS and GenCP)
are compared on constrained sentences generation tasks extracted from benchmarks recently introduced  \cite{yao2023collie}. The approach highlights how CP can provide guarantees when combined with LLM predictions.
% Theses benchmarks are also interesting because the authors shows that various LLMs struggle to produce length-constrained sentences.

The paper is organized as follows: Sec.~\ref{sec:prelim} serves as background, Sec.~\ref{sec:mmethod} shows how to extend \otf
to GenCP
and how to implement an interaction between 
% \otf and LLM. 
GenCP and LLM.
Sec.~\ref{sec:experiments} presents the experimental results  in which the new approach is demonstrated on the task of \tg. 
% In Sec.~\ref{sec:discussion}, the paper delves into further discussion, offering additional insights into this work and providing perspectives for future research endeavors. 
Finally, Sec.~\ref{sec:discussion} delves into further discussion, offering additional insights into this work and providing perspectives for future research endeavors.

\section{Background}
\label{sec:prelim}

This section introduces the necessary background on LLM and CP.

\subsection{LLM Predictions Strategies}

\subsubsection{Decoding Strategies Combined with LLMs}
Large Language Models (LLMs), such as the GPT series, generate text by predicting the next token (word or character) given the history of previously generated words. Decoding in LLMs refers to the strategy used to select the next words to be generated.

\subsubsection{Greedy Decoding}
The simplest decoding strategy is greedy decoding. Here, the LLM selects the words with the highest probability at each time step. Although simple and efficient, this approach does not guarantee the best overall sequence, as it does not consider the effect of the current selection on future tokens.

\subsubsection{Beam Search}

Beam Search (BS) \cite{beamsearch1:2021,beamsearch2:2018,hokamp-liu:2017} is a refined version of greedy decoding. A beam is a candidate sequence of words. Instead of selecting the single best token at each time step, it usually keeps track of the $k$ most likely sequences (beams) at each step.

Although BS usually achieves better results than greedy decoding, it assumes that high-ranking token sequences consist of high-ranking tokens, which may only sometimes be the case. For a more stochastic and diverse output, top-$k$ sampling and top-$p$ sampling (also known as nucleus sampling) are used. In top-$k$ sampling, the model selects from the top $k$ highest probability predictions, while in top-$p$ sampling, it dynamically selects the number of top predictions to cover $p$ percent of the total probability mass.

\subsubsection{Perplexity}
Perplexity is an entropy metric derived from Shannon's information theory \cite{brown-etal:1992}. 
Since an LLM computes the probability of text, then it can compute text perplexity.
It can be expressed as the geometric mean of the inverse conditional likelihood of the sequence \cite{jurafsky:2009}. Let $S_n$ be the sequence of a succession of words of size $n$: $S_n = w_1w_2..w_n$. The perplexity (PPL) of $S_n$ is computed as follows:
\[ PPL(S_n) = \sqrt[n]{\frac{1}{P(w_1w_2w_3...w_n)}},\]
where probability $P(\cdot)$ is given by the LLM. 
PPL can be interpreted as the ``how likely a text is generated by a given model'' \cite{garbacea-arxiv-survey-nlp:2022}.
Usually, it is used to evaluate the LLM itself by checking that good samples are recognized as such (i.e., low PPL values). 

In NLP, the evaluation of text is still an open problem, and human evaluation remains the gold standard. Numerous metrics have been developed to address this issue. Among them, PPL remains an objective criterion associated with text produced by a given model. PPL is also much more convenient to use than pure probability. Its range is $[ 1 ; + \infty [$ . The lower, the better.

\subsection{Constraint Programming} 

Constraint Programming (CP) is a paradigm for solving combinatorial problems that draws on a wide range of techniques from artificial intelligence and operations research. 
In CP a problem can be defined as a Constraint Satisfaction Problem (CSP). A CSP is a triplet: $\langle X, D, C \rangle$, where:
\begin{itemize}
\item $X = \{X_1, X_2, ..., X_n\}$ is the set of variables  of the problem.
\item $D = \{D_{X_1}, D_{X_2}, ..., D_{X_n}\}$ is the set of domains, where each domain $D_{X_i}$ corresponds to the set of possible values for the variable $X_i$.
\item $C = \{c_1, c_2, ..., c_m\}$ is the set of constraints of the problem. A constraints represent a property of the problem.
\end{itemize}
A solution is an assignment of all the variables to a value present in their respective domains, such that all the constraints are satisfied.

\subsubsection{Avoiding Static Definition of the CSP}

In traditional CP, for the task of \tg, a variable represents a word. Since the domains of variables have to be set beforehand, they will be of enormous size, containing every word/declination of words for a given language. 
Furthermore, constraints between succession of words may lead to a combinatorial explosion.
Since traditional CP is not well suited, this work focuses on \otf, a CP based method recently introduced by Régin and De Maria \cite{FGCSPICTAI}. Instead of having the variables/domains/constraints set before the search, \otf  generates the variables/domains/constraints during the search for a solution, 
avoiding the problem stated above and being expendable to permit the integration of an LLM.
The new version of \otf is called GenCP.

\section{Method: LLM alongside OTFS}
\label{sec:mmethod}
% \flo{Ici introduit \otf, explique amelioration c'est helping function et generation function, mets sur le graphe une bulle à cote du 2 avec LLM qui revient vers 2 pour indiquer apres (dans le text dira callLLM)}

% FGCSP formalism allows the \otf definition of a variable domain, permitting us to embed LLM in CP, combining CP and ML. Our new approach use a reformated FGCSP \cite{FGCSPICTAI} and embedded LLM in it.
%The approach is based on \otf \cite{FGCSPICTAI} (\otf), a variant of CSP that generates variables/domains/constraints during the search for a solution. 

\begin{figure}[h]
    \centering
    \includegraphics[width=0.80\textwidth]{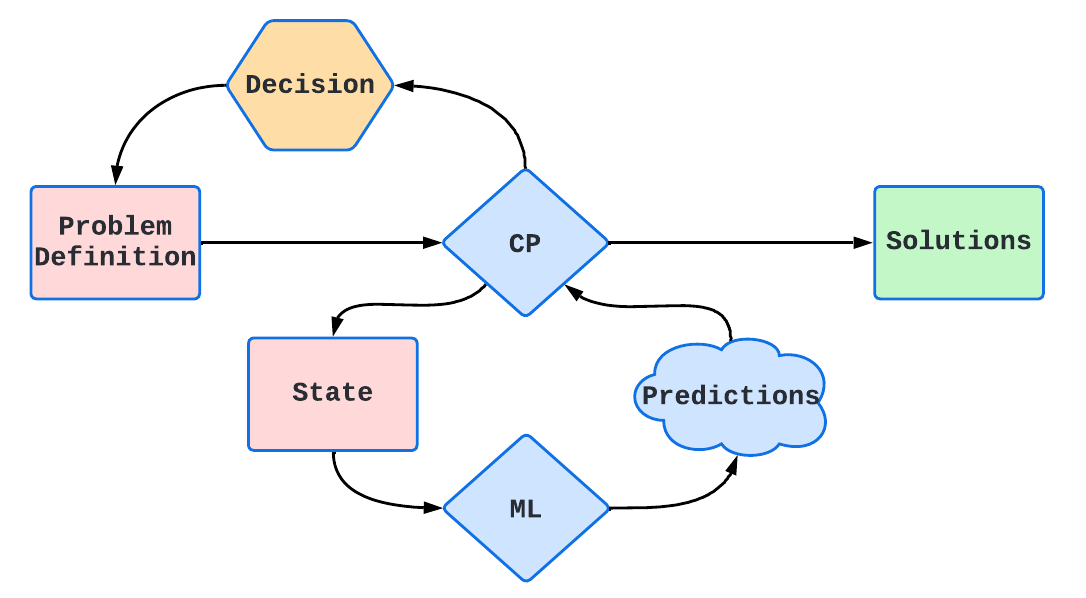}
     \caption{This scheme presents the integration of ML into CP performed by GenCP. It is freely inspired by Sec. 3.2.3 of Bengio et al.'s survey \cite{BENGIO-survey:2021}, which introduces an architecture for ML alongside Optimization Algorithms. The similarity is highlighted because the master algorithm (here, GenCP) repeatedly queries an ML model (here, an LLM) to obtain a prediction as a subroutine. In the context of this paper, the decision (search or propagation) has an impact on the problem definition (the CSP) because it may generate new variables, domains, or constraints during the solving process. The state is the current assignment of the variables.}
    \label{fig:toplevel}
\end{figure}

The approach of this paper extends \otf by having an embedded LLM generate the domains of variables. Figure \ref{fig:toplevel} graphically depicts the interplay between those components. The approach also adds a minor improvement in the form of helping functions, to differentiate between implicit constraints (prevent infinite loops, ensure a variable represents a word, etc.) and explicit constraints (constraints of the problem). 
% In the next subsection, the new version of \otf is described (from now on, \otf is directly used to refer to the modified version).
In the next subsection, the new version of \otf called GenCP is described.

\subsection{New version of \otf: GenCP}

In traditional CP it is not common to generate new variables/domains/constraints during the search, while \otf is based on this idea.
\otf begins with an empty or partially defined CSP (the CSP has less variables/domains/constraints than the CSP in traditional CP) and will generate variable/domains/constraints during the search for solutions.

GenCP is a new version of \otf that makes two changes to the original version: \textbf{1)} the function that generates the domain $genD$ calls an LLM to generate the domain of the current variable. \textbf{2)} Helping functions are added to represent implicit constraints.

Here is GenCP applied to \tg.
An GenCP model can be defined as a pair of sets $\{\mc{M},\mc{F}\}$, where:
\begin{itemize}
    \item $\mc{M} = \{X,D,C\}$ represents the model of the problem.
    \item $X$ represents the variables. The variables represent words.
    \item $D$ represents the domain of the variables. A domain $d_i \in D$ contains a list of predicted words by an LLM.
    \item $C$ represents the explicit constraints (constraints of the problem). A constraint $c_i \in C$ represents rules over text (e.g., number of words, number of characters, forbidden words, or symbols).
    \item $\mc{F} = \{G,B,H\}$ is a set of functions. 
    \item $G$ represents the set of generative functions: these functions explain to the solver how to generate variables/domains/constraints. 
    \item $B$ represents the set of Boolean functions: these functions tell the solver when a solution is found. 
     \item $H$ represents the set of helping functions: these functions are used to represent implicit constraints, for example ensuring that when a variable is generated, it helps obtaining a solution (to prevent the solver from attaining an infinite loop of generating variables).
\end{itemize}

\subsubsection{Generative Functions}

The set of generative functions $G = \{genV, genD, genC\}$ is such that:

\begin{itemize}
    \item $genV$ generates a new variable with an empty domain and adds it to $X$.
    \item $genD$ calls the LLM with the current sentence formed by the model and sets the domain of the previously generated variable to the output.
    \item $genC$ generates the constraint(s) relevant to the current variables of the model to $C$. The constraints generated depend on the problem (e.g., generate a sentence that does not contain the letter ``e"). 
\end{itemize}

\subsubsection{LLM integration}\label{sec:callLLM}

A variable is generated with an empty domain. To generate the domain of variables, $genD$ calls an LLM using $callLLM(sentence, parameters, k)$, where:

\begin{itemize}
    \item $sentence$ is the current sentence represented by the variables of the model.
    \item $parameters$ represents sampling parameters ($top\_k, top\_p$...). For this paper, $top\_k$ is used exclusively for both GenCP and BS: the LLM answers $k$ words ranked by probability, highest to lowest.
    \item $k$ is the number of words asked to the LLM.
\end{itemize}

Since the parameters and $k$ are not modified after the definition of the model, \\ $callLLM(sentence, parameters, k)$ will be simply referred to as $callLLM(sentence)$.

\subsubsection{Helping Functions}

Helping functions represent implicit constraints, like avoiding infinite loops.
In our current implementation, the set of functions $H$ contains the following functions:

\begin{itemize}
    \item $H_{o}$: it orders the domain of variables depending on the problem.
    \item $H_{onlyWords}$: it ensures that any word predicted by the LLM is a complete word and not the suffix or prefix of a word and it filters out any symbol or special character.
\end{itemize}

\begin{figure}
    \centering
    % \scalebox{0.95}{
    \scalebox{1}{
    \scriptsize
    % \begin{tikzpicture}[every node/.style={noeud, text width=12mm, align=center, inner sep=0pt}]
    \begin{tikzpicture}[every node/.style={noeud, align=center, inner sep=0pt}]

        \node[ellipse] (1) at (0, 0) {3: Helping\\Functions};

        \node[ellipse] (2) at (3, 0) {4: Save\\State};
        % \draw [arc] (1) edge[bend left] node[draw=none, above]{$\neq \emptyset$} (2);
        % \draw[arc] (1) edge[bend left] (2);
        \draw[arc] (1) edge[bend right] (2);

        \node[ellipse] (5) at (0, -1.5) {2: Generative\\Functions};
        % \draw [arc] (5) -- (2);
        \draw[arc] (5) -- (1);
        % \draw [arc] (1) -- node[draw=none, left]{$= \emptyset$} (5);

        \node[ellipse] (3) at (3, -1.5) {5: Run\\Propagation};
        \draw [arc] (2) -- (3);

        % \node[draw=none] (be) at (9, -2) {Propagation:\\remove values of variables\\that do not satisfy\\the constraints};

        \node[ellipse] (8) at (6, 0) {8: Backtrack};
        % \draw [arc] (8) edge[bend right] (2);
        \draw [arc] (8) edge[bend left] (2);
        \draw [arc] (3) edge[bend right] node[draw=none, below right]{fail} (8);
        % \draw [arc] (1.north west) edge[bend left] (8.north east);

        % \node[draw=none] (be) at (9, -3.9) {Backtrack:\\return to a previous\\saved state\\change};  

        \node[ellipse] (4) at (3, -3) {6: Boolean\\Functions};
        \draw [arc] (3) -- (4);
        % \draw[arc] (4) edge[bend left] node[draw=none, below]{no solution found} (5);
        % \draw [arc] (4) edge[bend left] (5);
        \draw [arc] (4) -- (5);
        % \node[ellipse,draw=none, rotate=-30] (999) at (1.5,-2.4) {no solution found};
        \node[ellipse,draw=none, rotate=-25] (666) at (1.9,-2.1) {no sol found};

        \node[ellipse] (6) at (6, -3) {7: Save\\Solution};
        % \draw[arc] (4) edge[bend right] node[draw=none]{solution found} (6);
        \draw[arc] (4) edge[bend right] (6);
        \node[ellipse,draw=none] (666) at (4.5,-3.75) {solution found};
        
        \draw [arc] (6) -- (8);

        \node[ellipse] (9) at (3, 1.5) {9: Return\\Solution(s) Saved};
        \node [ellipse, minimum width=33mm, minimum height=10mm] (9b) at (9) {};
        % \draw [arc] (8) edge[bend left] node[draw=none, above]{fail} (9b);
        \draw [arc] (8) edge[bend right] node[draw=none, above]{fail} (9b);

        % \node[ellipse] (10) at (1.5, 1) {Call Helping\\Functions};
        % \draw [arc] (5) -- (10);
        % \draw [arc] (10) -- (2);
        %  \draw[arc] (10) edge[bend left] node[draw=none]{fail} (8);

        \node[ellipse] (1b) at (0, -3) {1: Initial\\State};
        % \draw [arc] (1b) -- node[draw=none, left]{$= \emptyset$} (5);
        % \draw [arc] (1b) edge[bend left] (5);
        \draw [arc] (1b) -- node[draw=none, right]{empty} (5);
        \draw [arc] (1b) edge[bend left = 65] (1);
        % \draw[arc] (1b.north west) ++ (-0.4, 0.4) -- (1b);
        \draw[arc] (1b.south west) ++ (0, -0.4) -- (1b);

        \draw[arc] (1) edge[bend left=20] (8);
        \node[ellipse,draw=none] (333) at (1, 0.6) {fail};
        % \draw[->, thick, > = latex] (1.north) edge[bend left] (8.north east);

        \node[ellipse, color=blue] (llm) at (1.5, -1) {LLM};
        % \draw [arc] (5) -- (2);
        \draw[arc, color=blue] (llm)  edge[bend right] (5);
        \draw[arc, color=blue] (5)  edge[bend right] (llm);

    \end{tikzpicture}}
    \caption{This graph illustrates the main steps in GenCP solving.}
    \label{fig:fgcsp}
\end{figure}
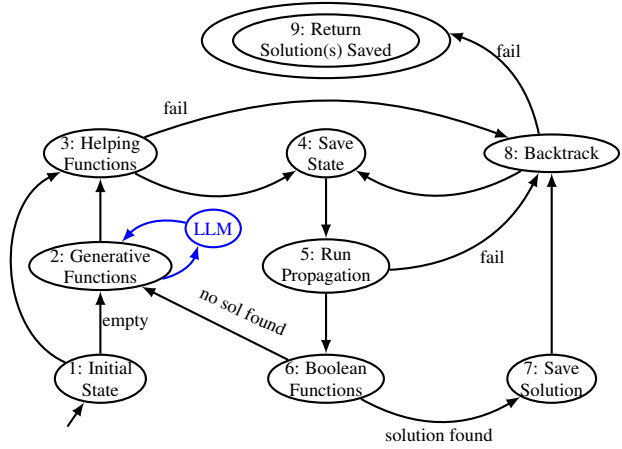

\begin{algorithm}[!ht]
\caption{GenCP($\mathcal{M}, \mc{F}$), $\mc{M} = \{X,D,C\}$, \mc{F} contains the generative and boolean functions}
\textbf{1:} $\mc{S} = \emptyset$; \textbf{if} $\mc{M}$ is not empty \textbf{then} go to \textbf{3.}\;
% \lIf{\mc{M} is not empty}{go to \textbf{3.}}
\textbf{2:} generativeFunctions($\mc{M}$)\;
\textbf{3:} helpingFunctions(\mc{M});
\textbf{if} $\mc{M}.$X$.containsEmptyVariable()$ \textbf{then} go to \textbf{8.}\;
% \lIf{\mc{M}.$X$.containsEmptyVariable()}{go to \textbf{8.}}
\textbf{4:} saveState($\mc{M}$)\;
\textbf{5:} propagation(\mc{M});
% \lIf{\mc{M}.$X$.containsEmptyVariable()}{go to \textbf{8.}}
\textbf{if} $\mc{M}.$X$.containsEmptyVariable()$ \textbf{then} go to \textbf{8.}\;
\textbf{6:} \lIf{not booleanFunctions(\mc{M})}{go to \textbf{2.}}
\textbf{7:} \mc{S}.add(\mc{M})\;
\textbf{8:} 
\textbf{if} $backtrack(\mc{M})$ \textbf{then} go to \textbf{4.}; \textbf{else} return \mc{S}\;
% \lIf{backtrack(\mc{M})}{go to \textbf{4.}} \lElse{return \mc{S}}

\label{alg:fgcsp}

\end{algorithm}

% $X_i$, $D_i$ and $C_i$ represent the sets of initial variables/domains/constraints. The sets can be empty.
% $\mathcal{M} = \{X = X_i, D = D_i, C = C_i, E\}$\;
% \tcp*{\textbf{1.}}

\subsubsection{Description of the new approach}

The main steps of GenCP are depicted in Fig. \ref{fig:fgcsp} and Algorithm \ref{alg:fgcsp}:

\begin{enumerate}
    \item GenCP begins with an initial state. If the initial state is empty, the generative functions are called (\textbf{2.}), otherwise the helping functions are called (\textbf{3.}).
    \item The generative functions $genV/genD/genC$ are called ($genD$ calls the LLM).
    \item The helping functions are called to manage implicit constraints, backtracking if necessary (e.g., if the LLM generated an empty domain).
    \item The current state of the model $\mc{M}$ is saved.
    \item The propagation is called, if it fails the model backtracks (8.), else it calls the boolean functions (6.).
    \item The Boolean functions are called to check if a solution has been found. If a solution is found, it is saved (7.) and the model backtracks (8.), otherwise the model calls the generative functions (2.).
    \item The current sentence formed by the variables is saved as a solution.
    \item GenCP backtracks to a previously saved state (4.) of the model and changes the choices made during propagation (5.). If no previous state was saved, then backtracking fails (9.).
    \begin{itemize}
        \item When backtracking to a previously saved state, the model deletes all variables, their respective domains, and the constraints associated with them, that are not present in the previously saved state.
    \end{itemize}
    \item GenCP outputs the solution(s) that were saved or it indicates that no solution was found.
\end{enumerate}

\subsubsection{Enforce variability}

Variability between two sentences is the number of words that are not equal at each position, for example:

\begin{itemize}
    \item ``The little boy is" and ``The little cat is'' have a variability of 1.
    \item ``My name is John'' and ``John is my name'' have a variability of 4.
\end{itemize}

To force a greater variability between solutions (greater than $2$), a special backtrack called $backtrackTo$($n$) is used. Let the set of variables $X = \{x_1,\dots,x_n,x_{n+1}\dots,x_m\}$. The function $backtrackTo$($n$) deletes the variables $x_{n+1}$ to $x_m$ and causes a backtrack.\label{bkTO}
For example,
     consider the sentence ``I like to swim in the summer.''. With $backtrackTo$(2), ``to swim in the summer.'' is deleted and the value of variable $x2 = ``like"$ is changed. The next solution might be ``I want to break free.''.
\subsubsection{Ordering}
\label{order}

For some tasks, not following the ordering strategies of the LLM (like top-$k$ and top-$p$) can lead to better/faster solutions.
Two other orderings are considered: PPL valuation and length of a word (depending on the average word length in the given language).

\subsection{Modeling Example}

Here is a simple example of how the search of GenCP works:
for this paper the generative functions only generate variables one at a time but it is important to note that these functions can generate multiple variables, domains and constraints at once.
Let us suppose GenCP has to generate a sentence beginning by ``The'' and containing between 10 and 15 words with exactly 60 characters.
The following functions are needed:
\begin{itemize}
    \item $currentSentence(\mc{M})$: outputs the current sentence the variables form.
    \item $callLLM(sentence)$: described in \ref{sec:callLLM}. Here $k$ is equal to 10 (each time the LLM is called, it will output 10 words).
    \item $contains(sentence, word)$: outputs yes if the sentence contains the word and no otherwise.
    \item $nbChar(sentence)$: outputs the number of characters in the sentence.
\end{itemize}
The obtained  model is $\{\mc{M},\mc{F}\}$, where:
\begin{itemize}
    \item $\mc{M} = \{X,D,C\}$:
     \item $X = \{x_1\}$.
    \item $D = \{d_1 = \{$``The''$\}\}$.
    \item $C = \emptyset$.
    \item $\mc{F} = \{G,B,H\}$.
    \item $G = \{genV, genD, genC\}$ is a set of functions, each function follows these steps:
    \begin{itemize}
        \item generate $x_{|X|+1}$ and add it to $X$ with an empty domain $d_{|X|+1}$.
        \item $d_{|X|+1} = callLLM(currentSentence(\mc{M}))$.
        \item $c_{remove_{over60char}((currentSentence(\mc{M}),d_{|X|+1})}$. 
        \item The constraints remove the words that make the current sentence exceed 60 characters from the domain of the current variable.
    \end{itemize}
    \item $B = \{endNbWords, endNbCharacters, endLLM\}$ is a set of functions, each function behaves as follows:
    \begin{itemize}
        \item $|X| >= 10 \land |X| <= 15$.
        \item $nbChar(currentSentence(\mc{M})) == 60$.
        \item $contains(callLLM(currentSentence(\mc{M})), ``.")$.
    \end{itemize}
    \item $H = \{H_{ho}\}$:
    \begin{itemize}
        \item $H_{ho} : order(d_{|X|+1})$.
        \item To help attain the goal of 60 characters, the domain of the current variable is ordered such that before the 10th word the solver tries the longer words first and at the 10th word the solver tries the shorter words first.
    \end{itemize}
\end{itemize}

With the above representation of the problem, GenCP is asked for 4 solutions, $backtrackTo(2)$ is used and the LLM is asked for 10 words maximum per call. The obtained solutions are:
\begin{enumerate}
   \item The following is an article by the author of the above book.
   \item The first time you see the movie version of your book on TV.
   \item The New York Times has an article on the new book by Tim Wu.
   \item The new year is here and we are ready to make the next step.
\end{enumerate}

\subsection{Illustrated Example}

\begin{figure*}[!ht]
    \centering
    \includegraphics[width=\textwidth]{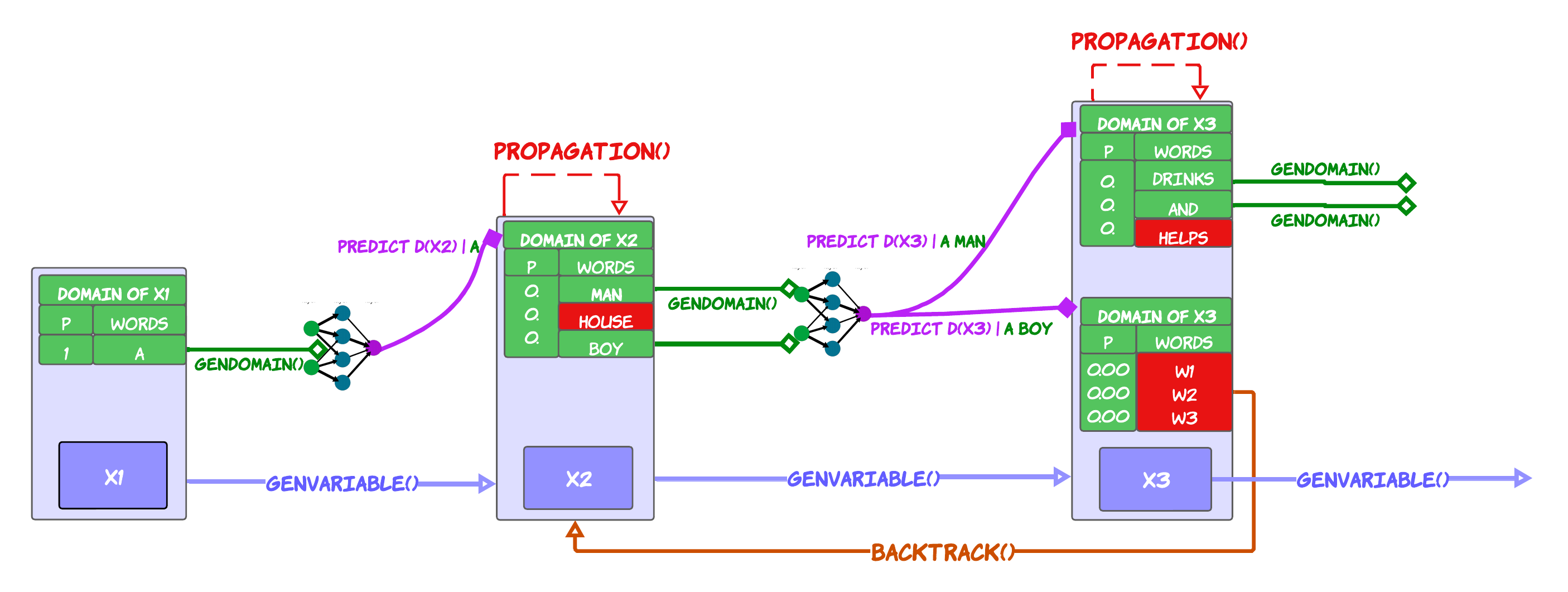}
    
     \caption{Illustrations of GenCP as a simplified framework with three variables and predictions of 3 values per LLM call, on a simple problem: generate a sentence that does not contain the letter \emph{e}. For each variable, the predefined constraint $c_i$ ``the letter \emph{e} is forbidden'' is generated. A predefined domain with one word is defined for the first variable: \color{darkgreen} A\color{black}. The current sentence formed by the variable ``A'' is not a solution ($callLLM(``A")$ does not answer a period (``.'')), so a new empty variable $x_2$ is generated. GenCP calls the LLM with the sentence ``A'' to predict the domain of $x_2$. The LLM model predicts three values: [~\color{darkgreen}man, house, boy\color{black}~]. $c_2$ is generated, causing the domain of $x_2$ to be filtered accordingly: \color{red} house \color{black} is removed. $x_2$ is then assigned to \color{darkgreen}boy\color{black}, GenCP generates the variable $x_3$ and calls the LLM with the sentence ``A boy'' to predict a new domain. Unfortunately, the domain of $x_3$ is empty, either because the LLM answered an empty domain or because this domain was entirely filtered during propagation. Hence, GenCP backtracks to $x_2$ and the value of $x_2$ is changed to \color{darkgreen}man\color{black}. In the same fashion as before, GenCP generates the variables $x_3$, and gives ``A man'' to the LLM that predicts: [~\color{darkgreen}drinks, and, helps\color{black}~]. $c_3$ is generated, filtering \color{red} helps \color{black} because it contains an \emph{e}. The process continues until the domain of the next predicted variable contains a period (a solution is found) or the solver \emph{fails}. }
    \label{fig:gensearch}
\end{figure*}

Fig. \ref{fig:gensearch} illustrates GenCP as a simplified framework with three variables and predictions of 3 values per LLM call, on a simple problem: generate a sentence that does not contain the letter \emph{e}.

\section{Experiments}
\label{sec:experiments}
\subsection{Experimental Conditions}

\subsubsection{Baseline}
\label{sec:baseline}
The experiments presented by Yao et al. are partially reproduced \cite{yao2023collie}.
In particular, the constrained sentence generation tasks described in Tab. \ref{tab:benchmarklist}. 
Five LLMs were selected: GPT4, GPT4-O, Mistral Next, Claude 3.5, and Gemini.
The four LLMs are prompted with the same example command given in \cite{yao2023collie}.
For example, ``Please create a sentence of exactly 82 characters.'' for the Sent-1 task\footnote{https://chatgpt.com/share/b2834735-f7d8-468a-ba54-7da19dd6723c}.
Tab. \ref{tab:iclrBaseline} gives an overview of the performance of the five LLMs on the four tasks. 
The satisfaction rate is based on ten trials per task per model. 
In addition, Tab. \ref{tab:iclrBaseline} also shows that the LLMs perform well on the lexically constrained scenario task-4 with a 90+\% satisfaction rate over ten trials.
Also, as Yao et al. previously showed in their paper, LLMs struggle to produce constrained sentences involving counting (e.g., words and characters). They provide a nice picture of current LLM satisfaction capabilities by introducing new benchmarks. Unfortunately, the Yao et al. article only provides the benchmarks and some hints on reproducing them. However, it does not give any hints on how to solve the tasks associated with the benchmarks (see the original article for more details~\cite{yao2023collie}).

\begin{table}[!ht]
    \centering
    \resizebox{!}{!}{ 
    \begin{tabular}{lrrr}
    \toprule
    name & words count & character count & lexical constraints  \\ 
    \midrule
     %sent-1& $\in [12,16]$ &  & is that a                \\
     %sent-2& $\in [18,22]$ &  & John, car seller, BMW \\
     sent-1&     & $=82$ &  \\
     sent-2& $=10$    &  & $X_3 =$ soft, $X_7 =$ soft,$X_{10} =$ math  \\
     
     sent-3& $\geq 20$   & $\forall i, |X_i| \le 6$ &   \\
    % sent-6&    &   & a, e, t as forbidden letters    \\
     sent-4&    &   & soft, beach, math   \\
     % \hline
    % sent-8&    & $\in [110,140]$, $\forall i, |X_i| \geq 4$  & beam search, constraint programming\\
     % & & & u is a forbidden letter\\
     \bottomrule
    \end{tabular}
    }
    \caption{
    Four tasks on sentence generation used to compare BS and GenCP extract from \cite{yao2023collie}.}
    \label{tab:benchmarklist}
\end{table}

\begin{table}[!ht]
    
    \centering
    \resizebox{\textwidth}{!}{
    \begin{tabular}{c|*{3}{c}|*{3}{c}|*{3}{c}|*{3}{c}|*{3}{c}}
    \toprule
    name & \multicolumn{3}{c}{GPT-4} & \multicolumn{3}{c}{GPT-4.0} & \multicolumn{3}{c}{Mistral Next} & \multicolumn{3}{c}{Claude3.5 Sonnet} & \multicolumn{3}{c}{Gemini}\\
      & \#s & \#f & \%sat & \#s & \#f & \%sat & \#s & \#f & \%sat & \#s& \#f & \%sat  & \#s& \#f & \%sat \\
     \midrule
     sent-1 &1&9& 10\%     &0&10& 0\%    &0&10& 0\%     &1&9& 10\%    &0&10& 0\%\\
     sent-2 &0&10& 0\%     &0&10& 0\%    &0&10& 0\%     &0&10& 0\%    &0&10& 0\%\\
     sent-3 &1&9& 10\%     &5&5& 50\%    &0&10& 0\%     &9&1& 90\%    &1&9& 10\%\\
     sent-4 &9&1& 90\%     &9&1& 90\%    &10&0& 100\%   &10&0&100\%   &1&9& 10\%\\
     \bottomrule
\end{tabular}
}
    \caption{Number of successes (\#s), Number of fails (\#f) and satisfaction rate (\%sat) for each model (GPT-4, GPT-4.0, Mistral Next, Claude 3.5, Gemini) for each task (sent-1, sent-2, sent-3, sent-4).}
    \label{tab:iclrBaseline}
\end{table}

\subsubsection{Hardware \& Implementation}
The experiments were performed on a laptop with Windows 10 Professional, 32 GB RAM, and Intel 16 CPU cores.
The approach and the BS are implemented in Java 17 for easier comparisons. 
\subsubsection{LLM choice}
 LLaMa~\cite{touvron2023llama} is responsible for the predictions of words as domains for the variables, mainly because an efficient implementation in C++ was recently released\footnote{https://github.com/ggerganov/llama.cpp}. It allows running a model on a personal laptop and CPU (only) efficiently. 
Thanks to quantization \cite{quantization_survey:2022} (model weight compression), the 7B parameters model (in Float16) of 13GB original size, in 4-bit quantization (Q4) drops to 3.9GB of RAM. However, the biggest model of LLama 65B (120GB), even in Q4, needs 38.5 GB of RAM. %
Thus, the LLaMa v1 model used in the experiments is LLaMa~7B~Q4 with low temperature (i.e., $\leq 1$, temp = 0.8). 

When asked for $k$ words, this version of LLaMa will take the same amount of time to ouput $1$ word and $1000$ words. To minimize the importance of $k$, $callLLM$ outputs more than $k$ words, a beam/variable only keeps $k$ ``valid" words. A ``valid" word is a word that does not violate a constraint on its own. For example, a word that does not violate the constraint ``does not contain the letter \emph{e}".

\subsubsection{\BS Technical Remarks}

In the current implementations two halting conditions are defined for BS:
\begin{itemize}
    \item First solution: when the current beam contains at least one solution, BS is stopped and output the solutions.
    \item All solutions: when the current beam contains at least one solution but another beam can continue to generate words without violating a constraint (for instance, it does not contain enough characters to satisfy a length constraint), the beam solutions are saved and BS continues with the remaining beams. %
\end{itemize}

\subsubsection{Benchmarks Settings}

BS and GenCP are compared  on some recent benchmarks described in Sec. \ref{sec:baseline}. %

To guarantee GenCP and BS to be judged on the generation of sentences of the same quality, a solution is a sentence that satisfies all the constraints of the current task and, when given this sentence, the LLM predicts a period (``."). Not to alter BS too much, words are ordered by probability (PPL is not used) and, since BS sentences have low variability, GenCP is used without $backtrackTo$($n$).

BS and GenCP are compared on the following criteria:
\begin{itemize}
    \item Time in seconds.
    \item Number of solutions. GenCP was stopped when it found the same number of solutions as BS on a task. $0/1$ means that BS found no solution while GenCP found one solution.
    \item The ratio solutions/outputs as a constraint satisfaction rate.
    \item For BS only, the number of bad outputs (number of outputs that are not a solution).
    \item For GenCP only, the number of backtracks.
\end{itemize}

\subsection{Result Analysis}

The results show that GenCP can be used to solve efficiently \tg problems. GenCP is faster than BS and all the outputs are solutions, contrary to BS where some outputs are not solutions.

Although the results suggest that GenCP succeeds in all tasks (see Tab. 3),
 it becomes particularly interesting when considering size constraints (e.g., sentences with a precise number of words or characters). It obtains sentences that satisfy the constraint with a low PPL score on sent-1 and sent-3 tasks.
 
GenCP also succeeds in producing sequences obeying lexical constraints in sent-2 and sent-4.
However, the PPL and a human evaluation on these sentences show a substantial deterioration in term of quality (i.e., meaningfulness).

Therefore, regarding sent-1 and sent-3 tasks, GenCP is to be preferred, whereas for sent-4 and sent-2 tasks, LLMs prompted alone or joint with BS is still adequate.

\begin{table}[htbp]
\centering
    \resizebox{!}{!}{
        \begin{tabular}{*{3}{c}|*{3}{c}|*{3}{c}}
        % \begin{tabular}{*{5}{|p{1.5cm}}||*{3}{p{1.5cm}|}}
            \toprule
            % \multicolumn{1}{c}{} & \multicolumn{7}{|c||}{Time (seconds)} & \multicolumn{7}{c|}{Space (megabytes)}\\
            % \hline
            % \N & \T& Temps CPU (s) & \# Backtrack & \# Nœuds & Temps CPU (s) & \# Backtrack & \# Nœuds & \# Appels utiles & \# Appels inutiles\\
            \multicolumn{3}{c}{Experiments} & \multicolumn{3}{c}{BS} & \multicolumn{3}{c}{GenCP}\\
            \midrule
            sent-i & k & \#sols & s & \#badoutput & \%sat & s & \%sat & \#bk\\
            \midrule

            1 & 5 & 1 & 108 & 9 & 10\% & 103 & 100\% & 45\\
             %\hline
             & 10 & 0 & 182 & 18 & 0\% &177 & 100\% & 84\\
             %\hline
             & 20 & 1 & 399 & 58 &1\%& 46 & 100\% & 13\\
             %\hline
             & 50 & 1 & 1123 & 109 & $\approx 0\%$ & 47 & 100\% & 13\\
            \midrule
            2 & 5 & 5 & 34 & 0 & 100\% &  38 & 100\% & 38\\
             %\hline
             & 10 & 10 & 69 & 0 &100\% & 36 & 100\% & 25\\
             %\hline
             & 20 & 20 & 140 & 0 &100\% & 58 & 100\% & 40\\
             %\hline
             & 50 & 49 & 354 & 1 & 99\% & 134 & 100\% & 100\\
            %\hline
            \midrule
            3 & 5 & 0 & 248 & 5 & 0\% & 36 & 100\% & 4\\
             %\hline
             & 10 & 2 & 510 & 8 & 20\% & 55 & 100\% & 6\\
             %\hline
             & 20 & 4 & 1030 & 16 & 20\% & 164 & 100\% & 38\\
             %\hline
             & 50 & 20 & 2633 & 30 & 66\% & 1174 & 100\% & 374\\
            %\hline
            \midrule
            4 & 5 & 25 & 279 & 3 & 89\%& 308 & 100\% & 118\\
             %\hline
             & 10 & 30 & 513 & 8 & 78\%  & 311 & 100\% & 114\\
             %\hline
             & 20 & 45 & 1123 & 14 & 76\% & 321 & 100\% & 104\\
             %\hline
             & 50 & 89 & 2928 & 40 & 68\% & 388 & 100\% & 27\\
            \bottomrule
        \end{tabular}
    }
    \caption{Comparison of BS and GenCP on the tasks of Tab. \ref{tab:benchmarklist}.
   Task considered (sent-i), Number of solutions (\#sols), Time in seconds (s), Number of bad output (\#badoutput), satisfaction rate (\%sat) and Number of backtracks (\#bk).  
    }\label{er:arrcpvbs}
\end{table}
\begin{table}[htbp]
\centering
    \resizebox{!}{!}{
        % \begin{tabular}{*{7}{|c}||*{3}{c|}}
        \begin{tabular}{*{1}{c}*{5}{c}}
        % \begin{tabular}{*{5}{|p{1.5cm}}||*{3}{p{1.5cm}|}}
            \toprule
            % \multicolumn{1}{c}{} & \multicolumn{7}{|c||}{Time (seconds)} & \multicolumn{7}{c|}{Space (megabytes)}\\
            % \hline
            % \N & \T& Temps CPU (s) & \# Backtrack & \# Nœuds & Temps CPU (s) & \# Backtrack & \# Nœuds & \# Appels utiles & \# Appels inutiles\\
            \multicolumn{3}{c}{Experiments} & \multicolumn{3}{c}{GenCP}\\
            \midrule
            sent-i & k & \#sols & s & MB & \#bk\\
            \midrule

            1 & 50 & 2 & 1123 & 136 & 79\\
             %\hline
            2 & 50 & 355 & 222 & 208 & 222\\
             %\hline
            3 & 50 & 488 & 2633 & 378 & 624\\
             %\hline
            4 & 50 & 830 & 2929 & 676 & 680\\
            \bottomrule
            
        \end{tabular}
    }
    \caption{GenCP results for $k=50$ when given approximately the same amount of time as BS in Tab \ref{er:arrcpvbs}. Number of solutions (\#sols), Time in seconds (s), Memory usage in megabytes (MB),  Number of backtracks (\#bk).  
    }\label{er:arrCPres}
\end{table}

% \color{black}
\subsubsection{\BS}

BS and GenCP are compared in Tab. \ref{er:arrcpvbs}. In all tables, the number of backtracks is denoted by \#bk. 
BS is slower than GenCP and has lower satisfaction rate (number of outputs that are solutions / total number of outputs), denoted by \%sat. This is due to multiple facts:

\begin{enumerate}
    \item \BS can not guarantee to find every solution.
    \item \BS chooses the next word depending on the probability of the LLM. \label{bs:proba}
    \item At each step, BS considers $k$ sentences, each sentence asks $k$ words to the LLM, so each step considers $k^2$ words. BS orders these words decreasingly by probability and only keeps the $k$ first. \label{bs:reduce}
\end{enumerate}

Facts \ref{bs:proba} and \ref{bs:reduce} explain why increasing $k$ does not guarantee to find the same/more solutions, it might even cause BS to find less solutions.

Let us suppose $k$ = 5, BS found one solution, and at depth 4, the candidate needed to find this solution was ranked 5 out of 25.
    Let us suppose now $k$ is increased to 6: at each step BS will consider 36 candidates and take the 6 best ones. BS considers 11 more candidates than with $k$ = 5; if at depth 4, the candidate needed to find the previous solution is now ranked 7 instead of 5, BS will not consider it and $k$ = 6 will not find the solution found with $k$ = 5.

\subsubsection{GenCP}

\begin{table}[htbp]
    \resizebox{\textwidth}{!}{
        % \begin{tabular}{*{7}{|c}||*{3}{c|}}
        %\begin{adjustbox}{angle=90}
        \begin{tabular}{*{1}{c}*{4}{r}}
        % \begin{tabular}{*{5}{|p{1.5cm}}||*{3}{p{1.5cm}|}}
            \toprule
            % \multicolumn{1}{c}{} & \multicolumn{7}{|c||}{Time (seconds)} & \multicolumn{7}{c|}{Space (megabytes)}\\
            % \hline
            % \N & \T& Temps CPU (s) & \# Backtrack & \# Nœuds & Temps CPU (s) & \# Backtrack & \# Nœuds & \# Appels utiles & \# Appels inutiles\\
            %\hline

            \multicolumn{2}{c}{Experiments} & \multicolumn{3}{r}{GenCP}\\
            \midrule
            sent-i & k & $bkTo$($n$) & PPL & sentence generated\\
            \midrule
            1 & 50 & NO & 8 & The following is an article by Dr David Hillon the   subject of the role of prayer.\\
             %\hline
             & & 2 & 13 &  The New York Times has an article on the  new book by former President George Bush.\\
            %\hline
             & & 3 & 6 & The following information is taken from the website of the National Park Services.\\
            \midrule
            2 & 50 & NO & 189 & The following soft skills are required beach resort jobs math.\\
             %\hline
             & & 2 & 169 &  The National soft drink association has beach balls and math.\\
            %\hline
             & & 3 & 107 & The most soft and comfortable of beach wear is math.\\
            \midrule
            3 & 50 & NO & 8 & The first time you see the movie The Big Short is like being hit by an ice cube in the face.\\
             %\hline
             & & 2 & 5 &  The world is full of great ideas and the best way to  get them out there is by using the power of the web.\\
            %\hline
             & & 3 & 5 & The first step in the right path is to know  what you want and where you are going in life.\\
            \midrule
            4* & 50 & NO & 347 & The following is an article by Dr math and  science teacher beach high school in soft.\\
             %\hline
             & & 2 & 593 & The term of the contract is for math and science teachers beach to be able soft.\\
            %\hline
             & & 3 & 48 & The following data is based on the math and  physics of beach waves and the soft sand.\\
            %\hline

            \bottomrule
            
        \end{tabular}
        %\end{adjustbox}
    }
    \caption{Output sentences of GenCP on the experiments of Tab \ref{tab:benchmarklist} associated with the task (sent-i), k, $backtrackTo$ ($bkTo$), and Perplexity (PPL). In sent-4* a constraint was added so that ``soft", ``beach", ``math" have to be separated by at least three words. 
    Sentences with high perplexity were chosen to showcase the importance of low perplexity.
    }\label{er:arrCPres2}
\end{table}

Tab. \ref{er:arrCPres} shows the capability of GenCP to generate more solutions than BS.
GenCP is given the same time as BS for the same task and $k=50$, GenCP obtains more solutions than BS. Note that for sent-1, without $backtrackTo$ GenCP only obtains 2 solutions in 1123 seconds, while with $backtrackTo(6)$ GenCP obtains 11 solutions in 1123 seconds.

The LLM-enhanced GenCP avoids the drawbacks of BS and proposes an alternative approach to \tg for the following reasons:

\begin{itemize}
    \item GenCP can guarantee to find every solution (if any). 
    Increasing $k$ guarantees to find at least the same solutions previously found and potentially finds new solutions. Furthermore, it can offer more solutions than BS.
    \item All the outputs answered by GenCP are solutions (all the constraints are satisfied). %
    \item GenCP offers more options for improvement, for example to ensure better variability ($backtrackTo$ explained in \ref{bkTO} can be used) or other orderings than probability (\ref{order}).
\end{itemize}

\subsubsection{Variability and Perplexity}

Tab. \ref{er:arrCPres2} demonstrates the importance of enforcing variability and perplexity.
When GenCP generated solutions for Tab. \ref{er:arrcpvbs} and \ref{er:arrCPres}, the maximum variability was 4. Tab. \ref{er:arrCPres2} shows that with $backtrackTo(2)$/$backtrackTo(3)$, sentences generated are almost completely different thanks to high variability (10+ for sent-3 for example).

Tab. \ref{er:arrCPres2} purposefully contains sentences with high perplexity to illustrate that this leads to a degradation in the sentence quality (i.e., low meaning).

All the sentences generated for sent-4 had the words ``soft", ``beach" and ``math" next to each other. To showcase the capability of GenCP to improve sentences, sent-4* was created: it is the same as sent-4 except that ``soft", ``beach" and ``math" must contain at least three words between them.

\section{Discussion \& Perspectives}
\label{sec:discussion}

\subsection{GPU and CPU Interplay}

The article shares a proof-of-concept showing that interesting results can be obtained using CPU resources combined with a small quantized LLM in a CP solver.
However, LLMs, in general, work best with much larger computational resources and require GPU resources.
Even though smaller models (e.g., Mistral 8x7B) sometimes manage to take top places in specific scenarios. The top spots in the LLM Elo rankings feature gigantic models \cite{chiang2024chatbot}. Given their size, clusters of GPU are quickly mandatory. 
Hence, it would be interesting to study in more detail how the joint use of resources (for instance, CPU for solver and GPU for LLM) could improve the results of the paper and correspond to more real-world usage in industry.

\subsection{Token Management}

In this article, GenCP ignores tokens and works at the word level (pre-token). 
It is possible to handle tokens by adapting the problem modeling. Indeed, it is possible to consider a word as a meta-variable $X_1$ composed of several decision variables (e.g., $X_{1_1}$, $X_{1_2}$, $X_{1_3}$...). This is useful and straightforward, as it is not clear in advance how the tokenizer will cut the words.
For instance, let us consider the following sentence: \emph{The first step in the recruitment of a new hire is to make sure that the job requisition is clear}.
Let us look at the assignments of the variables (space separates meta-variables, and semicolon decision variables): \emph{The; first; step; in; the; rec;ruit;ment; of; a; new; h;ire; is; to; make; sure; that; the; job; requ;is;ition; is; clear;}. The word \emph{recruitment} needs three decision variables because it is composed of three tokens (i.e., \emph{rec, ruit} and \emph{ment}).
It is easy to manage in GenCP because it can generate as many variables as required. Nevertheless, the evolution of the CSP (generation of variables and domains) is rather technical and, therefore, depends on the tokenizer.

\subsection{CSP Modeling}

The idea that a CSP can evolve in response to external information is not new (e.g., Dynamic Constraint Network \cite{dechter-dechter:88dcn}). This dynamic vision of CSPs has been motivated by several real-world problems, particularly in product configuration \cite{junker2006handbook}. GenCP proposes ML integration in modeling by letting LLMs manage operations for CSP domains during the resolution process. The "outside the world" information \cite{bessiere:91acindynamiccsp} is given by the LLM. The article shows that LLMs can contribute to CSP modeling for generation tasks. However, how ML/LLMs can be used for CSP modeling in general for any problem remains an open problem \cite{Freuder_2024,lawless2024iwantwayenabling,tsouros2023holygrail20natural,serdarkadioglu:2023ner}.

\section{Conclusion}
\label{sec:conc}

This paper showed that combining CP solving of structural constraints and ML understanding of vague notions (like meaning) on the task of \tg obtains promising results. 
This paper presents GenCP, a new method that extends \otf to make the domains manageable by LLM predictions.
The results show that GenCP can generate meaningful sentences that ensure various properties like the number of words, number of characters, mandatory keywords, or some forbidden characters. The results also show that GenCP has 100\% satisfaction rate and takes less time to output solutions of the same quality than a well-known technique in the field of text generation under constraints: Beam Search. 
GenCP provides multiple improvements thanks to ordering, enforcing variability and perplexity, allowing thus to obtain overall higher quality solutions than BS.

\section*{Acknowledgments}
We thank Jack Massey for his help in reproducing the benchmarks used as baseline in Section 4.1.1.

\bibliographystyle{unsrt}
\bibliography{biblio}

\end{document}